\documentclass[lettersize,journal]{IEEEtran}
\usepackage{amsmath,amsfonts}
\usepackage{amssymb} %
\usepackage{algorithmic}
\usepackage{algorithm}
\usepackage{array}

\usepackage{textcomp}
\usepackage{stfloats}
\usepackage{url}
\usepackage{verbatim}
\usepackage{graphicx}
\usepackage{cite}
\usepackage{xcolor} 
\usepackage{multirow}
\usepackage{makecell}
\usepackage{booktabs}

\ifCLASSOPTIONcompsoc
    \usepackage[caption=false, font=normalsize, labelfont=sf, textfont=sf]{subfig}
\else
\usepackage[caption=false, font=footnotesize]{subfig}
\fi

\def\ie{\emph{i.e}}

\begin{document}

\title{MoTe: Learning Motion-Text Diffusion Model for Multiple Generation Tasks}

\author{Yiming Wu, Wei Ji, Kecheng Zheng, Zicheng Wang, Dong Xu,~\IEEEmembership{Fellow,~IEEE}
\thanks{Yiming Wu, Zicheng Wang, and Dong Xu are with the Department of Computer Science, University of Hong Kong, Hong Kong. E-mail: yimingwu@hku.hk, xiaoyao3302@outlook.com, dongxu@hku.hk}
\thanks{Corresponding author: Dong Xu.}
\thanks{Wei Ji is with the School of Computer Science and Engineering, National University of Singapore, Singapore. E-mail: weiji0523@gmail.com}
\thanks{Kecheng Zheng is with the School of Computer Science and Technology, Zhejiang University. E-mail: zkechengzk@gmail.com.}}

\markboth{Journal of \LaTeX\ Class Files,~Vol.~14, No.~8, August~2021}%
{Shell \MakeLowercase{\textit{et al.}}: A Sample Article Using IEEEtran.cls for IEEE Journals}

\maketitle

\begin{abstract}
Recently, human motion analysis has experienced great improvement due to inspiring generative models such as the denoising diffusion model and large language model. While the existing approaches mainly focus on generating motions with textual descriptions and overlook the reciprocal task. In this paper, we present~\textbf{MoTe}, a unified multi-modal model that could handle diverse tasks by learning the marginal, conditional, and joint distributions of motion and text simultaneously. MoTe enables us to handle the paired text-motion generation, motion captioning, and text-driven motion generation by simply modifying the input context. Specifically, MoTe is composed of three components: Motion Encoder-Decoder (MED), Text Encoder-Decoder (TED), and Moti-on-Text Diffusion Model (MTDM). In particular, MED and TED are trained for extracting latent embeddings, and subsequently reconstructing the motion sequences and textual descriptions from the extracted embeddings, respectively. MTDM, on the other hand, performs an iterative denoising process on the input context to handle diverse tasks. Experimental results on the benchmark datasets demonstrate the superior performance of our proposed method on text-to-motion generation and competitive performance on motion captioning.

\end{abstract}

\section{Introduction}
\label{sec:intro}

AI-powered generative models have captured global attention in recent years, with prominent software such as Midjourney and Gen-2 demonstrating the ability to generate counterfeit images and videos with the given prompt. Not limited to image~\cite{rombach2022high} and 3D content generation~\cite{poole2023dreamfusion}, human motion synthesis has also experienced great improvement due to the inspiring denoising diffusion model (DM)~\cite{ho2020denosing,song19generative} and large language model (LLM)~\cite{radford2021learning}. Emerging research takes a huge leap forward regarding diversity and fidelity, which should benefit the development of games, film production, and the AR/VR industry.

\begin{figure*}[t]
    \centering
    \includegraphics[width=0.95\linewidth]{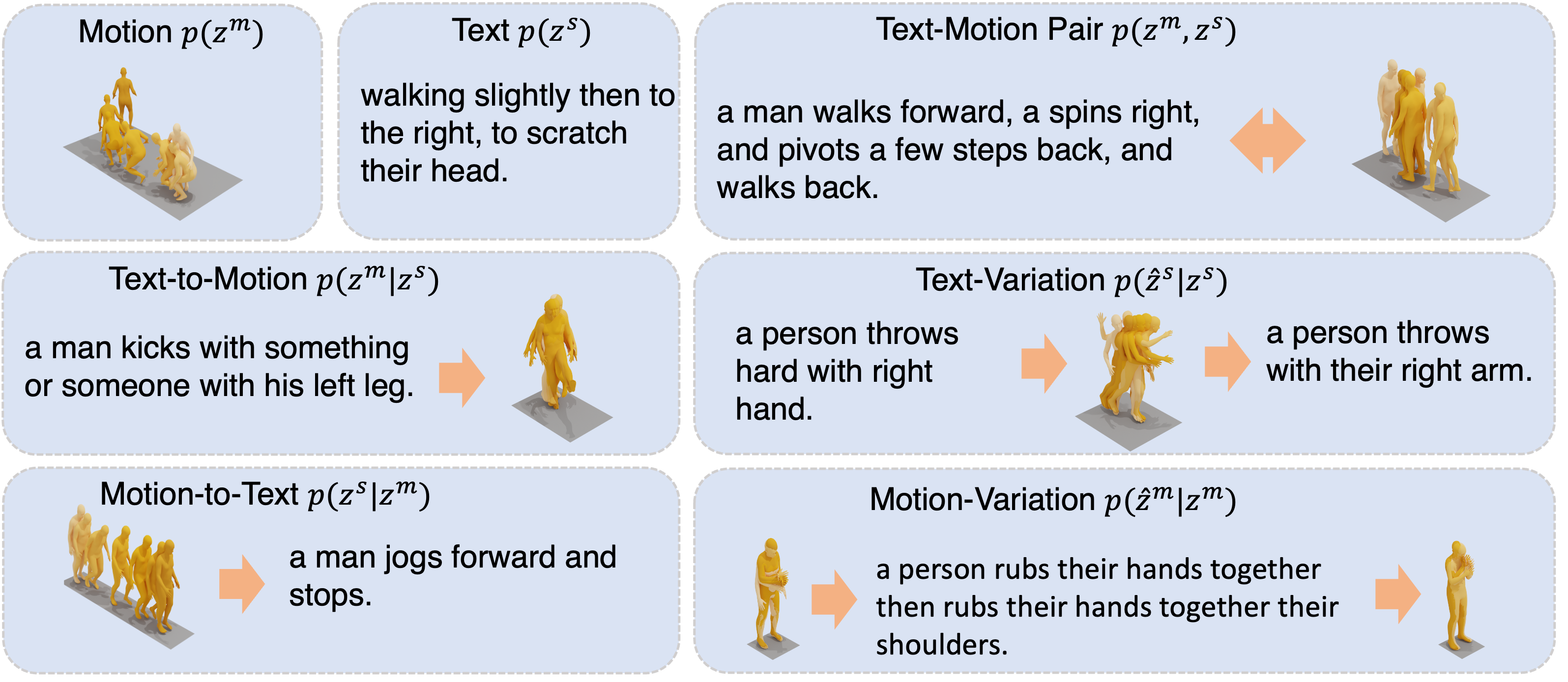}
    \caption{Illustration of diverse tasks: (a) The first row presents the unconditional and joint generation results. Motion, text, and text-motion pairs are generated from Gaussian noise. (b) The second row and third rows indicate conditional generation. Giving text or motion as input, MoTe can perform different tasks like text-to-motion and motion-to-text, as well as variation.} %
    \label{fig:multitask}
\end{figure*}

Due to the user-friendly nature and flexibility of language description, there has been extensive exploration into motion generation based on textual descriptions, which is known as text-based motion generation (text-to-motion)~\cite{tevet2022human,zhang2022motiondiffuse,chen2023executing,zhang2023t2mgpt}. Most of the current text-driven motion generation approaches perform the generation process in the latent feature space, which shows great potential. However, these methods are restricted to generating content for a single modality, limiting their adaptability to diverse tasks. Nowadays, multi-modality capability is considered the next path of achieving artificial general intelligence. However, only a few works like TM2T~\cite{guo2022tm2t} and MotionGPT~\cite{jiang2023motiongpt} consider the multi-modal generation task for human motion synthesis, while the motion generation performance of these approaches is far from satisfying. MotionGPT, for instance, treats motion tokens the same as word tokens, neglecting the inherent distribution differences between latent motion embeddings and latent text embeddings. In contrast, we employ modality-specific transformer layers with an interaction module in our proposed diffusion model to align motion and text.

To build up the multi-modal generative model for diverse motion-related tasks, it is critical to correctly align motion and language modalities. Despite the redundancy and noise present in raw motion sequences and language descriptions, motion and language share similarities in the semantic space. Therefore, a viable solution is to align these two modalities in the latent space. Additionally, the ability to perform multitask generation corresponds to learning the marginal, conditional, and joint distribution within a single model. To this end, we present \textbf{MoTe} (\textbf{Mo}tion-\textbf{Te}xt Diffusion Model) for multiple motion-related tasks. Specifically, our proposed method operates in a two-stage manner. In the first stage, we train the motion encoder-decoder (MED) and text encoder-decoder (TED) for two modalities, and then the motion sequences and language descriptions are mapped to latent spaces, respectively. In the second stage, a diffusion model is optimized to generate latent motion and text embeddings from noise. During the diffusion model training stage, we find that the interaction module between motion and language features is critical for learning the distributions, and we provide an analysis of different interaction modules. Moreover, our observations indicate that simply adopting the condition generation loss during training could accelerate convergence. We further explore the impact of motion embedding on the performance of text-to-motion and motion-to-text tasks, the experimental results indicate a trade-off between these two tasks. Furthermore, with only a simple modification of the input context, MoTe demonstrates capability in performing various tasks, including motion generation, text generation, and variations.

In summary, our contributions can be summarized in three main aspects: 1) We present ~\textbf{MoTe}, a motion-text diffusion model that is proficient in handling multiple tasks, including motion generation (\ie, random motion, text-to-motion), text generation (\ie, random text, motion-to-text), and the variations (\ie, text-variation, motion-variation), with these different tasks solvable through minor modifications of the input context. We hope this simple method serves as a new baseline and provides the community with a new perspective;  2) We provide a detailed analysis for variants of interaction modules in the proposed method, determining that the In-Context interaction module performs better than the alternative modules in most cases, and we adopt the In-Context interaction module in MoTe; 3) We evaluate our method on two widely used datasets, and observe a trade-off between motion-to-text and text-to-motion tasks. Compared with the previous work, our method achieves the best performance in most metrics for the text-to-motion task and achieves competitive performance for the motion-to-text task.

\section{Related Work}\label{sec:related}
\noindent\textbf{Human Motion Synthesis.} The field of human motion synthesis involves the creation of human-like movements using various inputs, such as action label~\cite{tevet2022human,petrovich2021action}, text~\cite{guo2022generating,chen2023executing,jiang2023motiongpt,zhang2023t2mgpt}, and audio~\cite{li2021ai}. Among these inputs, text is particularly explored due to its user-friendly nature and flexibility. Prior works have employed RNN~\cite{plappert2018learning}, VAE~\cite{petrovich2022temos}, and GANs~\cite{goutsu2021linguistic} to tackle this challenging task. Notably, recent methods based on the diffusion model~\cite{ho2020denosing} and GPT~\cite{radford2021learning} have demonstrated remarkable performance due to their exceptional fitting capabilities. MDM~\cite{tevet2022human} for the first time employs a diffusion model for diverse motion-related tasks, including action-to-motion, text-to-motion, motion in-between, and motion prediction. Subsequently, MLD~\cite{chen2023executing} advances the latent diffusion model for generating human motion conditioned on various inputs. Fg-T2M~\cite{wang2023fg} emphasizes the significance of fine-grained language descriptions and introduces a multi-step progressive inference strategy within the existing diffusion model framework. Moreover, T2M-GPT explores the application of GPT to human motion generation, proposing a framework based on VQVAE and GPT, and achieving state-of-the-art performance on text-to-motion. On the other hand, research endeavors in motion captioning remain constrained. In~\cite{plappert2018learning,yamada2018paired}, motion and text features are separately extracted using two autoencoders, enabling text and motion generation from shared latent vectors. Additionally, sequence-to-sequence RNN is adopted in~\cite{yamada2018paired} to translate motion to text and vice versa. SeqGAN~\cite{goutsu2021linguistic} extends the neural machine translators (NMT) model with a discriminator for text-to-motion and motion-to-text. And TM2T~\cite{guo2022tm2t} employs two separate neural machine translators for bi-modal mutual mappings between 3D human motions and texts. A recent contribution, MotionGPT~\cite{jiang2023motiongpt}, extends Flan-T5~\cite{chung2022scaling} to address multiple motion-related tasks with instruction tuning.

\noindent\textbf{Multi-modal and Multi-task Generation Model.}
The pivotal role of multi-modal and multi-task generative models in universal AI is underscored by recent advances in vision and natural language processing. The diffusion model~\cite{ho2020denosing, song19generative}, LLM~\cite{radford2021learning, touvron2023llama}, and various foundation models~\cite{cai2023smpler} have opened new horizons. To fulfill the multi-modal content generation, the first step is understanding multi-modal content involves learning multi-modal features through vision-and-language pretraining (VLP). CLIP~\cite{radford2021learning} is the first VLP trained on the large-scale dataset with a contrastive loss that maximizes the likelihood of matched image-text pairs and minimizes mismatched pairs. BEIT-3~\cite{wang2022image} achieves state-of-the-art performance through masked modeling on image, text, and image-text pairs. ImageBind~\cite{girdhar2023imagebind} extends the multi-modal capability to six modalities by aligning all modalities with image data. Building upon these foundation models, the diffusion model and LLM are scaled from single modality to multi-modality. MM-Diffusion~\cite{ruan2023mmdiffusion} for the first time generates synchronized audio and video using a multi-modal diffusion model. Versatile Diffusion (VD) combines multiple diffusion pipelines into a unified model, capable of handling text-to-image, image-to-text, and their variants simultaneously. UniDiffuser~\cite{bao2023one} designs a single diffusion model that fits all distributions with minimal modification to the input context, achieving competitive performance with bespoke text-to-image diffusion models. Beyond the diffusion model, LLM extends its generation capability by integrating an image generator with a language model. Flamingo~\cite{alayrac2022flamingo}, BLIP-2~\cite{li2023blip}, and LLaVA~\cite{liu2023visual} introduce a learnable interface between the pretrained visual encoder and LLM. A most related work in human motion synthesis is MotionGPT~\cite{jiang2023motiongpt}, which establishes a uniform motion-language framework on top of the pretrained LLM, unlocking the versatility of the multi-modal large language model with instruction tuning. In contrast, our work explores a multi-modal diffusion motion model to handle motion-related tasks in a unified manner.

\section{Methodology}\label{sec:method}

\begin{figure*}[t]
    \centering
    \includegraphics[width=1\linewidth]{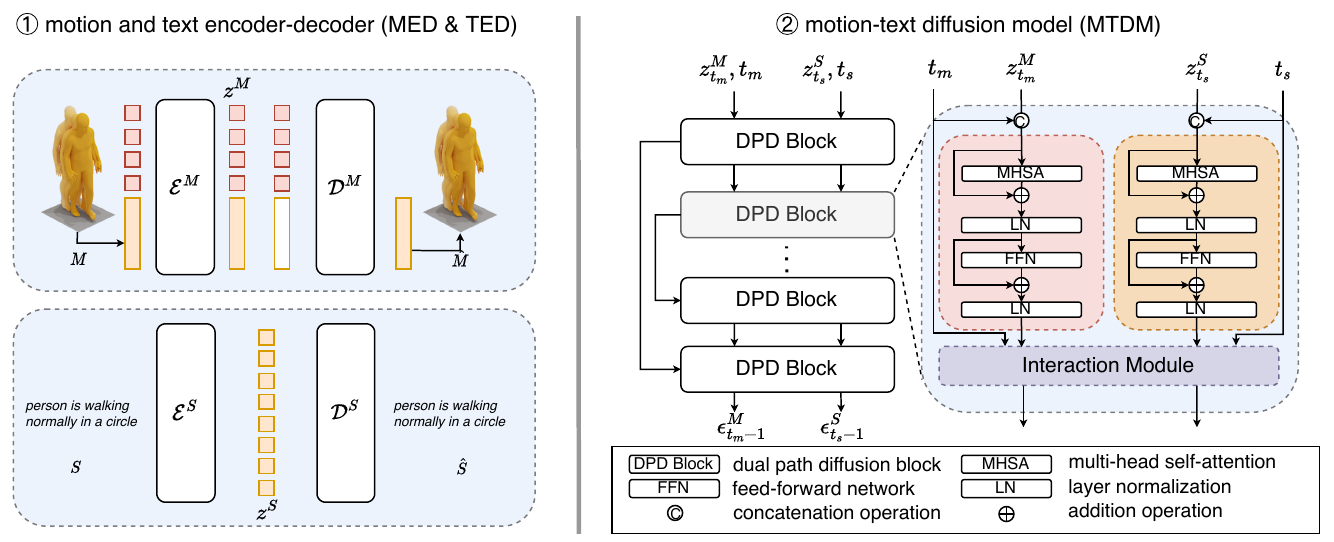}
    \caption{The overview of our proposed~\textbf{MoTe}. (1) Motion encoder-decoder (MED) and text encoder-decoder (TED) compress the motion sequences and language descriptions into two latent representations, which are reconstructed by the corresponding decoders. We adopt Motion Transformer in MED and CLIP-GPT2 in TED. (2) Motion-text diffusion model (MTDM) maps Gaussian noise through stacked dual path diffusion (DPD) blocks, where each DPD block consists of two unimodal transformer blocks and an interaction module.}
    \label{fig:framework}
\end{figure*}

In this section, we present our proposed multi-modal motion diffusion model (\ie, MoTe) designed to address various motion-related tasks. Before elaborating on the specific design of our method, we first revisit the preliminary work of the diffusion model, motion encoder-decoder (MED), and text encoder-decoder (TED) in Sec.~\ref{sec:preliminary}. Then, we present the specific design of MTDM in Sec.~\ref{sec:mtdm}. After that, we demonstrate how to perform multiple tasks with MoTe in Sec.~\ref{sec:multitask}. We briefly illustrate the overview of our proposed method in Fig.~\ref{fig:framework}.

\subsection{Preliminary}\label{sec:preliminary}
\noindent{\textbf{Diffusion Model.}} Diffusion model~\cite{ho2020denosing,song19generative} involves a diffusion process and a denoising process. In the diffusion process, the original data $z_0 \sim q(z)$ are perturbed through the addition of small-scale noise (Gaussian noise in general), the diffusion process is formulated as follows:
\begin{equation}
    \begin{aligned}
        q(z_t|z_{t-1}) &= \mathcal{N}(z_t | \sqrt{\alpha_t} z_{t-1}, (1-\alpha_t) \mathrm{I}), \\
        q(z_{1:T}|z_0) &= \prod_{t=1}^T q(z_t|z_{t-1}),
    \end{aligned}
\end{equation}
where $t\in[1,2,\dots, T]$ is the discrete timestep, and $\alpha_{1:T}$ is the pre-defined noise scheduler. The denoising process reverses the diffusion process by approximating the noise added in the $t$-th step:
\begin{equation}
    \begin{aligned}
       & p(z_{t-1}|z_t) = \mathcal{N}(z_{t-1} | \mu_t(z_t, t), \sigma_t^2 \mathrm{I}), \\
       & \mu_t(z_t, t) = \frac{1}{\sqrt{\alpha_t}} \big( z_t - \frac{1 - \alpha_t}{\sqrt{1 - \bar{\alpha}_t}} \epsilon_\theta(z_t, t) \big),
    \end{aligned}
\end{equation}
where $\bar{\alpha}_t = \prod_{i=1}^t \alpha_i$, and $\mu_t(z_t, t)$ denotes the estimated mean value of $z_{t-1}$ predicted by a denoising network $\epsilon_\theta$, and the variance $\sigma_t$ remains generally preserved during the denoising process, owing to its minor impact~\cite{nichol2021improved} on enhancing the fidelity of generated data. The denoising neural network $\epsilon_\theta$ is trained by maximizing the log-likelihood $\mathbb{E}[{\log p(z_0)}]$, which is simplified as a weighted mean squared loss:
\begin{equation}
    \min_{\theta} \mathbb{E}_{\epsilon \sim \mathcal{N}(0, \mathrm{I})} || \epsilon^z - \epsilon_{\theta}(z_t, t)||_2^2.
\end{equation}

For condition-based generation, the condition $c$ is injected into the network $\epsilon_{\theta}$, and the network models the conditioned distribution $p(z_t|c)$ by optimizing 
\begin{equation}
    \min_{\theta} \mathbb{E}_{\epsilon \sim \mathcal{N}(0, \mathrm{I})} || \epsilon^z - \epsilon_{\theta}(z_t, c, t)||_2^2.
\end{equation}

\noindent{\textbf{Latent Motion Embedding.}}~\label{sec:latent motion embedding} The raw motion representation, denoted as a motion sequence $M$, consists of joint velocities, positions, foot contact, and rotations. However, due to its inherent redundancy and noise, it is inefficient and difficult to reconstruct the raw motion representation for the diffusion model~\cite{chen2023executing}, we instead employ a motion encoder to obtain the latent motion embedding. Specifically, the motion encoder $\mathcal{E}^{M}$ encodes the motion sequence as a latent motion embedding $z^m = \mathcal{E}^M(M)$, and the motion decoder $\mathcal{D}^{M}$ reconstructs the raw motion sequence $\hat{\mathrm{M}}=\mathcal{D}^M(z^m)$.

Referring to the existing method, two types of motion encoder-decoder were proposed in the previous methods. In MotionCLIP~\cite{tevet22motionclip} and MLD~\cite{chen2023executing}, the motion sequences are encoded into a fixed-length latent embedding. In contrast, T2M-GPT~\cite{zhang2023t2mgpt} proposes a VQVAE based on 1D convolution layers, with the latent motion embedding being temporally downsampled by 4.

\noindent{\textbf{Latent Text Embedding.}}~\label{sec:latent text embedding} As for the language description, the raw text could be also represented as the token sequence $S$. For the purpose of generating the texts, we follow~\cite{bao2023one} to utilize CLIP and GPT2. Specifically, we first encode the text with CLIP~\cite{radford2021learning} and an additional linear layer, latent text embedding $z^s = \mathcal{E}^S(S)$ is then feed into the decoder $\mathcal{D}^S$ to reconstruct the original text $\hat{S} = \mathcal{D}^S(z^s)$.

\subsection{Motion-Text Diffusion Model}~\label{sec:mtdm}
In the pursuit of handling multitasks with a unified diffusion model, we extend the denosing network from a single modality (\ie, $\epsilon_{\theta}(z_t, t)$) to the multi-modality (\ie, $\epsilon_{\theta}(z^m_{t_m}, t_m, z^s_{t_s}, t_s)$), where $z^m$ and $z^s$ are latent embeddings for motion and language, $t_m$ and $t_s$ are the timesteps for motion and language, respectively. With this formulation, we could describe multiple data distributions in a single framework, such as $\epsilon_{\theta}(z^m_{t_m}, t_m, \varnothing, 0)$ and $\epsilon_{\theta}(\varnothing, 0, z^s_{t_s}, t_s)$ for marginal distribution $q(z^m_0)$ and $q(z^s_0)$; $\epsilon_{\theta}(z^m_{t_m}, t_m, z^s_0, 0)$ and $\epsilon_{\theta}(z^m_0, 0, z^s_{t_s}, t_s)$ for conditional distribution $q(z^m_0|z^s_0)$ and $q(z^s_0|z^m_0)$; and $\epsilon_{\theta}(z^m_{t_m}, t, z^s_{t_s}, t)$ for joint distribution $q(z^m_0,z^s_0)$.

Using the latent motion embedding $z^m_0$ and latent text embedding $z^s_0$, we could obtain the perturbed embeddings $z^m_{1:T}$ and $z^s_{1:T}$ by iteratively adding Gaussian noise. The denoising network $\epsilon_{\theta}$ are trained by minimizing the reweighed objective $\mathcal{L}_{joint}$ as follows:
\begin{equation}
   \mathcal{L}_{joint} = \mathbb{E}_{\epsilon^{z^m, z^s} \sim \mathcal{N}(0, \mathrm{I})} || \epsilon^{z^m, z^s} - \epsilon_{\theta}(z^m_{t_m}, t_m, z^s_{t_s}, t_s)||_2^2.
\end{equation}
To further enhance the optimization, we optimize the conditional generation loss:
\begin{equation}
    \begin{aligned}
        \mathcal{L}_{cond} = &\mathbb{E}_{\epsilon^{z^m} \sim \mathcal{N}(0, \mathrm{I})} || \epsilon^{z^m} - \epsilon_{\theta}(z^m_{t_m}, t_m, \varnothing, 0)||_2^2 + \\
        &\mathbb{E}_{\epsilon^{z^s} \sim \mathcal{N}(0, \mathrm{I})} || \epsilon^{z^s} - \epsilon_{\theta}(\varnothing, 0, z^s_{t_s}, t_s)||_2^2. 
    \end{aligned}   
\end{equation}
Therefore, the final training objective is $\mathcal{L} = \mathcal{L}_{joint} + \mathcal{L}_{cond}$.

\begin{figure}[t]
    \centering
    \scalebox{1.0}{
    \includegraphics[width=1\linewidth]{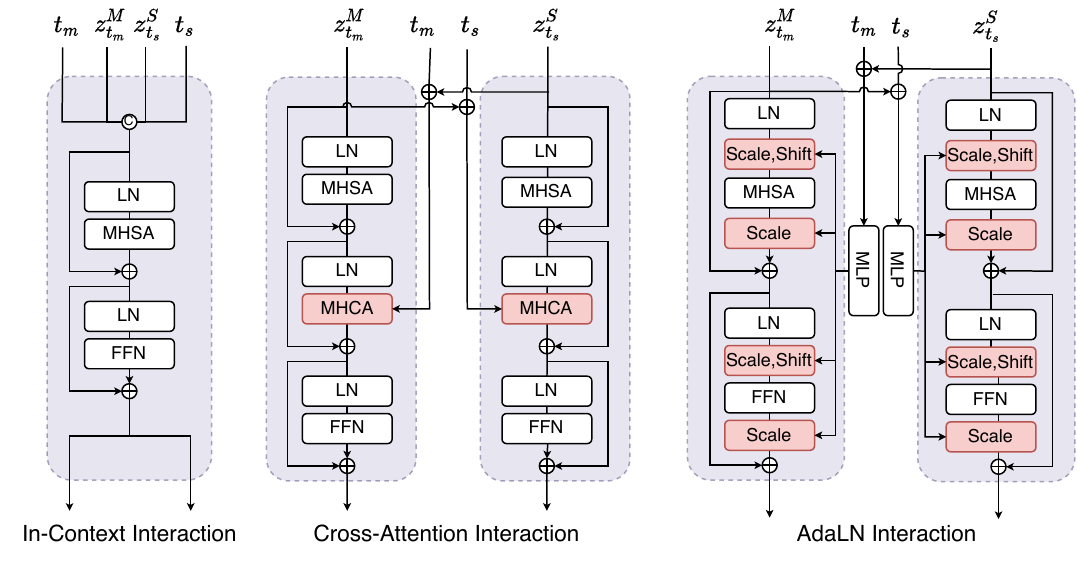}
    }
    \caption{Three variants of the interaction module: 1) In-Context Interaction, all embeddings are concatenated and then processed by the vanilla transformer block. 2) Cross-Attention Interaction, a multi-head cross-attention layer is inserted for modality interaction. 3) AdaLN Interaction, features are modulated by multiple scale-shift operations with the corresponding weights of which is generated by feeding the summation of timestep embedding and modality feature.}
    \label{fig:interaction module}
\end{figure}

To capture the multiple distributions in a diffusion model, designing an effective interaction module is critical. Previous works~\cite{rombach2022high,peebles2023scalable} have demonstrated the effectiveness of transformer block and UNet for generating uni-modal data content. Therefore, our proposed MoTe, illustrated in Fig.~\ref{fig:framework}, adopts a UNet structure comprising multiple dual path diffusion (DPD) blocks. In each DPD block, the motion embedding and textual embedding are first fed into the unimodal transformer block (post-normalization transformer block), and the outputs are subsequently processed in the multi-modal interaction module. As shown in Fig.~\ref{fig:interaction module}, we explore three variants of transformer-based multi-modal interaction modules: In-Context Interaction, Cross-Attention Interaction, and AdaLN Interaction. For simplicity, we introduce the $z^m$ generation process in the following interaction module since the interaction module is symmetric:
\begin{itemize}
    \item In-Context Interaction. Similar to the module used in MLD~\cite{chen2023executing}, all embeddings $t_m$, $z^m$, $t_s$, $z^s$ are concatenated and then processed by a vanilla transformer block, the embeddings interact in the multihead self-attention layer, at the final DPD block, the tokens for motion and text are kept while the timestep tokens are dropped.
    \item Cross-Attention Interaction. For processing motion embedding $z^m$, similar to the architecture of transformer decoder~\cite{vaswani2017attention}, a multi-head cross-attention layer is inserted between the multihead self-attention layer and the feedforward layer. Timestep $t_m$ and textual embedding $z^s$ are treated as the key and value in the cross-attention layer, and motion embedding $z^m$ serves as the query feature.
    \item AdaLN Interaction. The adaptive normalization layer, known for its success in GANs~\cite{karras2021astyle} and DiT~\cite{peebles2023scalable} is employed in this module, where $\alpha$, $\beta$, and $\gamma$ are generated by feeding the summation of timestep embedding $t_m$ and modality feature $z^s$ into an MLP layer. Subsequently, the motion embedding $z^m$ is modulated by scaling and shift operations. Following DiT~\cite{peebles2023scalable}, the weight of the MLP is initialized to zero to make the scale-shift operation the identity function initially.
\end{itemize}

\subsection{Motion-Text Generation}~\label{sec:multitask}
In our proposed multi-modal diffusion model, all distributions are modeled, enabling us to handle multiple motion-related tasks within a single framework by simply modifying the input context. The description of multiple motion-related tasks in this unified framework is as follows:
\begin{itemize}
    \item Unconditional Generation: To initiate unconditional generation, we set the condition as $\varnothing$ and perform iterative inference on the unimodality, in this case, $\epsilon_{\theta}(z^m_{t_m}, t_m, \varnothing, T)$ and $\epsilon_{\theta}(\varnothing, T, z^s_{t_s}, t_s)$ correspond to the noise predicted in the $t_m$-th and $t_s$-th steps, respectively. Here, $T$ represents the maximum diffusion step defined in the noise scheduler.
    \item Conditional Generation: For the conditional generation, the given condition and its corresponding timestep remain fixed during the denoising process. $\epsilon_{\theta}(z^m_{t_m}, t_m, z^s_0, 0)$ predicts latent motion noise added in the $t_m$-th step with the given textual description $z^s_0$. Similarly, $\epsilon_{\theta}(z^m_0, 0, z^s_{t_s}, t_s)$ predicts the latent text noise conditioned on motion.
    \item Joint Motion-Text Generation: For this task, $z^m$ and $z^s$ are both denoised from the $T$-th step, with the timesteps for motion and language empirically shared. This corresponds to $\epsilon_{\theta}(z^m_{t_m}, t, z^s_{t_s}, t)$.
\end{itemize}

Classifier-free guidance~\cite{ho2021classifier} is widely adopted to improve the diversity of conditionally generated content, we employ it in our training and inference stage. For conditional and joint motion-text generation, the estimated noise is calculated by:
\begin{equation}
    \begin{aligned}
        \epsilon_{\theta}^M &= w_m\epsilon_{\theta}(z^m_{t_m}, t_m, z^s_0, 0) + (1-w_m)\epsilon_{\theta}(z^m_{t_m}, t_m, \varnothing, 0),\\
        \epsilon_{\theta}^S &= w_s\epsilon_{\theta}(z^m_0, 0, z^s_{t_s}, t_s) + (1-w_s)\epsilon_{\theta}(\varnothing, 0, z^s_{t_s}, t_s),
    \end{aligned}
\end{equation}
where $w_m$ and $w_s$ are the scalars for classifier-free guidance, $\epsilon_{\theta}(z^m_{t_m}, t_m, \varnothing, 0)$ and $\epsilon_{\theta}(\varnothing, 0, z^s_{t_s}, t_s)$ are estimated noise without condition.%

\section{Experiment}
\subsection{Datasets}\label{sec:dataset}
We evaluate our proposed method on two datasets: HumanML3D and KIT. 

\begin{itemize}
    \item \textbf{HumanML3D}. This is a large 3D human motion-language dataset curated from HumanAct and AMASS datasets. It comprises 14,616 motions and 44,970 language descriptions, consisting of 5,371 distinct words in total. The motions are extracted from 28.59 hours of videos and sub-sampled to 20 fps, the motion length is 7.1 seconds on average. Besides, the average length of language descriptions is 12 words.
    \item \textbf{KIT}. This dataset contains 3,911 motions and 6,278 language descriptions, totaling 52,903 words. The video duration is 11.23 hours, and the motion sequences are sampled at 12.5 fps. Each motion is annotated with up to four language descriptions.
\end{itemize}
Following the previous works~\cite{chen2023executing,guo2022tm2t,jiang2023motiongpt,zhang2023t2mgpt}, we adopt HumanML3D format representation in our method, each motion feature is represented by joint velocities, positions, rotations, and foot contact. More details about motion representation are presented in the supplement.

\subsection{Evaluation Metrics}\label{sec:evaluation metrics}
For the assessment of our proposed method, we focus on text-to-motion and motion-to-text tasks.
For the text-to-motion task, we follow the previous works ~\cite{chen2023executing,guo2022tm2t,jiang2023motiongpt,zhang2023t2mgpt} to evaluate motion quality, diversity, and text-motion matching score. There are five metrics in total: 1) Retrieval-Precision (R-Precision). For each generated motion, groundtruth motion and 31 randomly mismatched motions are first sampled from the gallery. With a pretrained feature extractor~\cite{guo2022tm2t}, motions are extracted as 512-dim global features. The gallery motion features are ranked by Euclidean distances, and Top-1/2/3 retrieval accuracy is reported as R-Precision for the generated motion. 2) Matching Score (MMDist). It measures the matching score of language description and the corresponding generated motions by calculating the feature distance between text and motion. 3) Fréchet Inception Distance (FID).  It computes the feature distribution divergence between generated motions and groundtruth motions. 
4) Multi-Modality (MModality). MModality calculates the variance of motion features with the same language description. 
For the motion-to-text task, we follow MotionGPT~\cite{jiang2023motiongpt} to evaluate text quality and motion-text matching score. Except for R-Precision and MMDist, linguistic metrics Bleu~\cite{papineni2002bleu}, Rouge~\cite{lin2004rouge}, CIDEr~\cite{vedantam2015cider}, and BertScore~\cite{zhang2020bertscore} are measured to assess the quality of generated textual descriptions. %

\subsection{Implementation Details}\label{sec:implementation}
For training MED, we employ Transformer-based Motion VAE introduced in MLD~\cite{chen2023executing}, the encoder and decoder consist of 3 layers with skip connection, the latent dimension is $l \times 768$, where $l=4$ in our MoTe model. For training TED, we adopt the CLIP-GPT2 model introduced in Unidiffuser~\cite{jiang2023motiongpt} which is composed of CLIP-ViT-L/14 text encoder and GPT2 text decoder. To enhance text generation quality, we finetune the GPT2 text decoder on the HumanML3D and KIT datasets. Additional details about text decoders are available in the supplement. For training  MTDM, consistent hyper-parameters are applied to both HumanML3D and KIT datasets. The AdamW optimizer with an initial learning rate of 1e-4, a weight decay of 0.01, a mini-batch size of 128, and a cosine learning rate scheduler~\cite{loshchilov2017sgdr} over 6,000 training epochs is employed. The architecture of MTDM includes an input dimension of 768, a hidden dimension of 1024 for the feedforward layer, and a dropout layer with a drop rate of 0.1. For the denoising process, we adopt DDIMScheduler, and the inference step is set to 100 for all experiments. The MoTe model is trained on two RTX3090 GPUs.

\subsection{Comparison with the State-of-the-Art Methods}\label{sec:the state-of-the-art}
\begin{table*}[t]
    \centering
    \normalsize
    \caption{Comparison with the the state-of-the-art methods for text-to-motion task on the HumanML3D and KIT datasets. We follow T2M~\cite{guo2022generating} to calculate these metrics. All the results are evaluated by running the experiments 20 times and the confidence interval of 95\% is considered. \textbf{Bold} and \underline{underline} indicate the best and the second-best result, respectively. The results are sorted by FID.}
    \scalebox{1}{
    \begin{tabular}{l l c c c c c c} 
        \toprule
        \multirow{2}{*}{Dataset} & \multirow{2}{*}{Methods} & \multicolumn{3}{c}{R-Precision $\uparrow$} & \multirow{2}{*}{FID $\downarrow$} & \multirow{2}{*}{MMDist $\downarrow$} & \multirow{2}{*}{MModality $\uparrow$} \\ 
        \cline{3-5}
        & & Top1 & Top2 & Top3 &  &  &  \\ 
        \hline
        \multirow{10}{*}{HumanML3D} 
        & TM2T~\cite{guo2022tm2t}&  $0.424^{\pm.003}$&  $0.618^{\pm.003}$&  $0.729^{\pm.002}$&  $1.501^{\pm.017}$&  $3.467^{\pm.011}$& $\underline{2.424}^{\pm.093}$\\ 
        & MotionDiffuse~\cite{zhang2022motiondiffuse}& $0.491^{\pm.001}$&  $0.681^{\pm.001}$&  $0.782^{\pm.001}$&  $0.630^{\pm.001}$&  $3.113^{\pm.001}$&  $1.553^{\pm.042}$\\ 
        & MDM~\cite{tevet2022human}&  $0.320^{\pm.005}$&  $0.498^{\pm.004}$&  $0.611^{\pm.007}$&  $0.544^{\pm.044}$&  $5.566^{\pm.027}$&  $\mathbf{2.799}^{\pm.072}$\\ 
        & MLD~\cite{chen2023executing}&  $0.481^{\pm.003}$&  $0.673^{\pm.003}$&  $0.772^{\pm.002}$&  $0.473^{\pm.013}$&  $3.196^{\pm.010}$&  $2.413^{\pm.079}$\\ 
        & Fg-T2M~\cite{wang2023fg}&  $0.492^{\pm0.03}$&  $0.683^{\pm.003}$&  $0.783^{\pm.002}$&  $0.243^{\pm.019}$&   $3.109^{\pm.007}$& $1.614^{\pm.049}$\\ 
        & MotionGPT~\cite{jiang2023motiongpt} &  $0.492^{\pm.003}$&  $0.681^{\pm.003}$&  $0.778^{\pm.002}$&  $0.232^{\pm.008}$&  $3.096^{\pm.008}$&  $2.008^{\pm.084}$\\ 
        & HuTuMotion~\cite{han2024hutumotion}&  $0.500^{\pm.003}$&  $0.686^{\pm.002}$&  $0.782^{\pm.002}$&  $0.224^{\pm.006}$&  $3.058^{\pm.009}$&  $0.966^{\pm.046}$\\
        & T2M-GPT~\cite{zhang2023t2mgpt}&  $0.491^{\pm.003}$&  $0.680^{\pm.003}$&  $0.775^{\pm.002}$&  $0.116^{\pm.004}$&  $3.118^{\pm.011}$&  $1.856^{\pm.011}$\\
        & ReMoDiffuse~\cite{zhang2023remodiffuse}&  $ 0.510^{\pm.005}$&  $0.698^{\pm.006}$&  $0.795^{\pm.004}$&  $0.103^{\pm.004}$&  $2.974^{\pm.016}$&  $1.795^{\pm.043}$\\
        & DLP~\cite{cai2024digital} & $0.495^{\pm.003}$&  $0.651^{\pm.004}$&  $0.792^{\pm.004}$&  $\underline{0.071}^{\pm.002}$&  $3.561^{\pm.017}$&  $0.452^{\pm.012}$\\
        & MoMask~\cite{guo2024momask}&  $\underline{0.521}^{\pm.002}$&  $\underline{0.713}^{\pm.002}$&  $\underline{0.807}^{\pm.002}$&  $\mathbf{0.045}^{\pm.002}$&  $\underline{2.958}^{\pm.008}$&  $1.241^{\pm.040}$\\
        \hline
        & MoTe (Ours) & $\mathbf{0.548}^{\pm.002}$ & $\mathbf{0.737}^{\pm.002}$ & $\mathbf{0.825}^{\pm.002}$ & $0.075^{\pm.004}$ & $\mathbf{2.867}^{\pm.012}$ & $2.399^{\pm.075}$	\\\hline\hline
        \multirow{10}{*}{KIT} 
        & TM2T~\cite{guo2022tm2t}&  $0.280^{\pm.006}$&  $0.463^{\pm.007}$&  $0.587^{\pm.005}$&  $3.599^{\pm.051}$&  $4.591^{\pm.019}$&  $\mathbf{3.292}^{\pm.034}$\\ 
        & MotionDiffuse~\cite{zhang2022motiondiffuse}& $0.417^{\pm.004}$&  $0.621^{\pm.004}$&  $0.739^{\pm.004}$&  $1.954^{\pm.062}$&  ${2.958}^{\pm.005}$&  $0.730^{\pm.013}$\\ 
        & Fg-T2M~\cite{wang2023fg}&  ${0.418}^{\pm.005}$&  $0.626^{\pm.004}$&  ${0.745}^{\pm.004}$&  $0.571^{\pm.047}$&   $ 3.114^{\pm.015}$& $1.019^{\pm.029}$\\ 
        & T2M-GPT~\cite{zhang2023t2mgpt}&  $0.416^{\pm.006}$&  $0.627^{\pm.006}$&  ${0.745}^{\pm.006}$&  $0.514^{\pm.029}$&  ${3.007}^{\pm.023}$&  $1.570^{\pm.039}$\\ 
        & MotionGPT~\cite{jiang2023motiongpt} &  $0.366^{\pm.005}$&  $0.558^{\pm.004}$&  $0.680^{\pm.005}$&  $0.510^{\pm.016}$&  $3.527^{\pm.021}$&  $2.328^{\pm.117}$\\ 
        & MDM~\cite{tevet2022human}&  $0.164^{\pm.004}$&  $0.291^{\pm.004}$&  $0.396^{\pm.004}$&  $0.497^{\pm.021}$&  $9.191^{\pm.022}$&  $1.907^{\pm.214}$\\ 
        & MLD~\cite{chen2023executing}&  $0.390^{\pm.008}$&  $0.609^{\pm.008}$&  $0.734^{\pm.004}$&  $0.404^{\pm.027}$&  $3.204^{\pm.027}$&  $2.192^{\pm.071}$\\ 
        & MoMask~\cite{guo2024momask}&  $\mathbf{0.433}^{\pm.007}$&  $\mathbf{0.656}^{\pm.005}$&  $\mathbf{0.781}^{\pm.005}$&  ${0.204}^{\pm.011}$&  $\mathbf{2.779}^{\pm.022}$&  $1.131^{\pm.043}$\\
        & HuTuMotion~\cite{han2024hutumotion}&  $0.409^{\pm.004}$&  ${0.640}^{\pm.004}$&  $\underline{0.766}^{\pm.005}$&  $\underline{0.201}^{\pm.064}$&  $3.082^{\pm.025}$&  $0.901^{\pm.035}$\\
        & ReMoDiffuse~\cite{zhang2023remodiffuse}&  $\underline{0.427}^{\pm.014}$&  $\underline{0.641}^{\pm.004}$&  ${0.765}^{\pm.055}$&  $\mathbf{0.155}^{\pm.006}$&  $\underline{2.814}^{\pm.012}$&  $1.239^{\pm.028}$\\
        \hline
        & MoTe (Ours)  & $0.419^{\pm.005}$& ${0.627}^{\pm.005}$& $0.741^{\pm.004}$& $0.256^{\pm.018}$& $3.216^{\pm.036}$& $\underline{2.615}^{\pm.103}$\\
        \bottomrule
    \end{tabular}
    }
\label{tab:the state-of-the-art text-to-motion humanml3d}
\end{table*}

\begin{table*}
    \centering
    \caption{Comparison with the the state-of-the-art methods for the motion-to-text task on the HumanML3D dataset. In alignment with the evaluation protocol proposed in~\cite{jiang2023motiongpt}, pre-processing is not applied to the groundtruth language descriptions.}%
    \scalebox{1}{
    \begin{tabular}{l c c c c c c c c}
        \toprule
        \multirow{2}{*}{Methods} & \multicolumn{2}{c}{R-Precision$\uparrow$} &\multirow{2}{*}{MMDist$\downarrow$} & \multirow{2}{*}{Bleu@1 $\uparrow$} & \multirow{2}{*}{Bleu@4 $\uparrow$} &  \multirow{2}{*}{Rouge $\uparrow$} &  \multirow{2}{*}{CIDEr $\uparrow$} & \multirow{2}{*}{BertScore $\uparrow$} \\ \cline{2-3}
        &  Top1 & Top3 &  &  &  &  &  & \\ \hline
        Real &  $0.523$&  $0.828$&  $2.901$ & -  & - & - & - & -\\ \hline
        TM2T~\cite{guo2022tm2t}&  $0.516$ &  $0.823$ &  $2.935$ &  $\mathbf{48.9}$ &  $7.00$ &  $\underline{38.1}$& $16.8$ & $\underline{32.2}$\\
        MotionGPT~\cite{jiang2023motiongpt}&  $0.543$ &  $0.827$ & $2.821$ &  $48.2$&  $\mathbf{12.47}$& $37.4$& $29.2$& $\mathbf{32.4}$\\\hline
        MTDM ($l$=2) & $0.559$ & $0.850$ & $2.710$ & $45.1$ & $10.47$ &$36.9$ &$29.0$ &$29.4$ \\
        MTDM ($l$=4) & $\underline{0.577}$& $\mathbf{0.871}$& $2.649$& $46.7$ & $11.15$ &$37.4$ &$\underline{31.5}$&$30.3$ \\
        MTDM ($l$=6) & $\mathbf{0.578}$ & $0.855$ & $2.651$ & $45.4$ & $10.33$ &$37.0$ &$29.1$ &$29.1$ \\
        MTDM ($l$=8) & $0.575$ & $\underline{0.860}$& $\underline{2.652}$& $\underline{48.3}$& $\underline{11.76}$ &$\mathbf{38.4}$&$\mathbf{32.3}$&$31.3$ \\
        \bottomrule
    \end{tabular}
    }
    \label{tab:the state-of-the-art motion-to-text humanml3d}
\end{table*}

As mentioned in Sec.~\ref{sec:multitask}, our proposed MoTe excels in handling various motion-related tasks. We present quantitative results for text-to-motion (\ref{tab:the state-of-the-art text-to-motion humanml3d}) and motion-to-text generation tasks (\ref{tab:the state-of-the-art motion-to-text humanml3d}). Additionally, we provide qualitative comparison results on the HumanML3D dataset on unconditional generation, conditional generation, and variation tasks.

\noindent\textbf{Comparison on Text-to-Motion Generation Task.} The text-to-motion generation task involves creating a motion sequence based on a given textual description. We assess our proposed MoTe against other state-of-the-art (the state-of-the-art) approaches including T2M~\cite{guo2022generating}, TM2T~\cite{guo2022tm2t}, Motiondiffuse~\cite{zhang2022motiondiffuse}, MDM~\cite{tevet2022human}, MLD~\cite{chen2023executing}, T2M-GPT~\cite{zhang2023t2mgpt}, Fg-T2M~\cite{wang2023fg}, MotionGP-T~\cite{jiang2023motiongpt}, HuTuMotion~\cite{han2024hutumotion}, MoMask~\cite{guo2024momask}, and DLP~\cite{cai2024digital} on both HumanML3D and KIT datasets. Adhering to the recommended evaluation protocol~\cite{guo2022generating}, all results undergo 20 times and are reported under a 95\% confidence interval. As presented in Tab.~\ref{tab:the state-of-the-art text-to-motion humanml3d}, the experimental results on HumanML3D demonstrate that MoTe achieves superior performance across all metrics except for MModality, which reveals the generated motion sequence has lower diversity with the given language descriptions. On the KIT dataset, our proposed MoTe achieves comparable results with HuTuMotion~\cite{han2024hutumotion}.

\noindent\textbf{Comparison on Motion-to-Text Generation Task.} Motion-to-text generation, also known as motion captioning, entails generating human-like language descriptions for each motion sequence. In alignment with the evaluation protocol proposed in~\cite{jiang2023motiongpt}, no pre-processing is applied to the groundtruth language descriptions. In Tab.~\ref{tab:the state-of-the-art motion-to-text humanml3d}, TM2T~\cite{guo2022tm2t} and MotionGPT~\cite{jiang2023motiongpt} are compared with our proposed MoTe, and the experimental results demonstrate the competitive performance of MoTe across most metrics.

\begin{figure*}[t]
    \centering
    \subfloat[\label{fig:user study}]{%
        \includegraphics[width=0.33\textwidth]{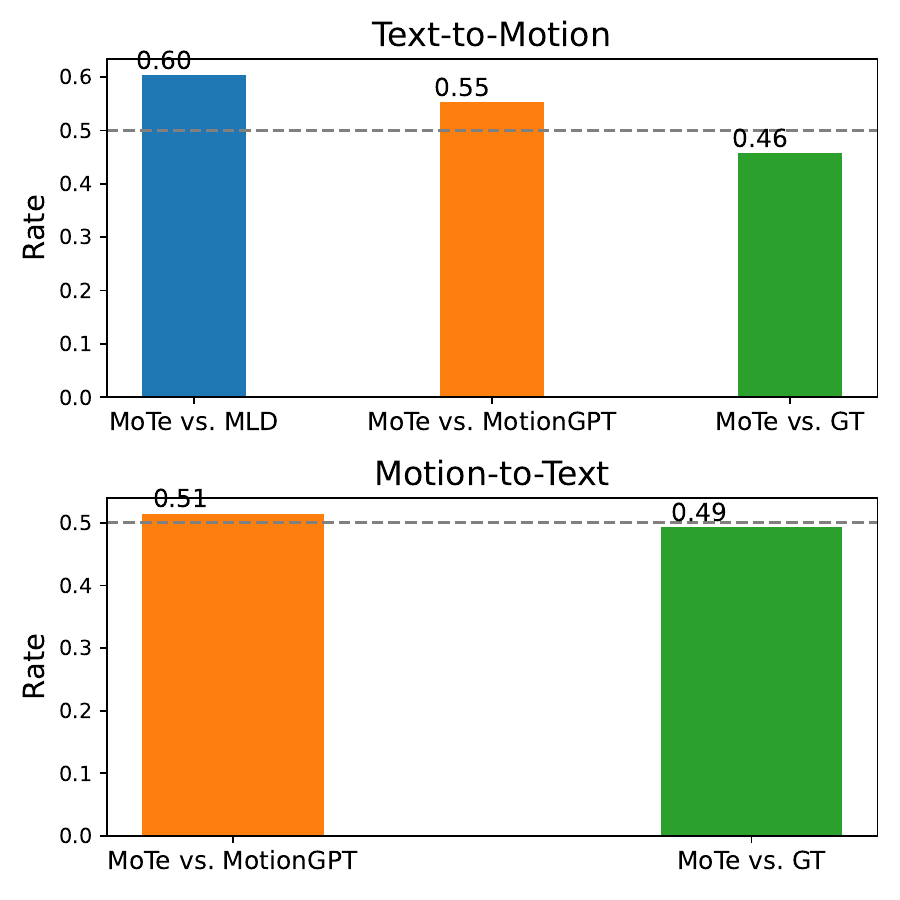}}
    \hfill
    \subfloat[\label{fig:qualitative}]{%
        \includegraphics[width=0.66\linewidth]{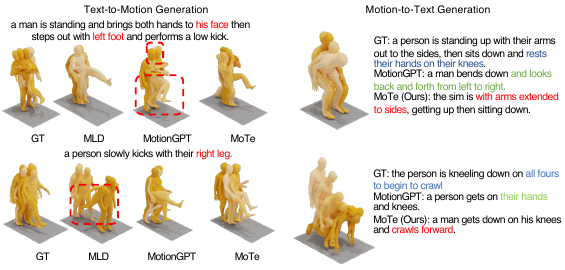}}   
    \caption{(a) Statistics of our user study for evaluating the text-to-motion and motion-to-text tasks. (b) Qualitative comparison on the HumanML3D dataset. We compare our proposed method MoTe with MLD~\cite{chen2023executing} and MotionGPT~\cite{jiang2023motiongpt} on the text-to-motion task, and compare our method with MotionGPT on the motion-to-text task.}
\end{figure*}

\noindent\textbf{User Study.} Except for the quantitative comparison based on the pre-trained feature extractor, we further present the statistics of our user study for evaluating the text-to-motion and motion-to-text tasks in Fig.~\ref{fig:user study}. For the text-to-motion task, the motions are generated from 20 textual descriptions from the HumanML3D test dataset. Our proposed method, MoTe, is systematically compared against two state-of-the-art approaches, namely MLD and MotionGPT. Similarly, for the motion-to-text task, we compare our proposed method with MotionGPT and collect the survey from 15 users. From the statistical analysis, our proposed MoTe achieves superior performance on the text-to-motion task, and it is competitive with MotionGPT and groundtruth on the motion-to-text task.

\noindent\textbf{Qualitative Comparison.} 
In~\ref{fig:qualitative}, we present visual results on the HumanML3D dataset. We compare our method with state-of-the-art methods MLD ~\cite{chen2023executing} and MotionGPT~\cite{jiang2023motiongpt}. From the rendered motion sequences, we find our model generates better human motion than the others with the given text descriptions. As for motion captioning, we compare our method with MotionGPT~\cite{jiang2023motiongpt}, the generated textual descriptions match the human motion. We provide more visual results in the supplement.

\subsection{Ablation Studies}\label{sec:ablation studies}
Our model undergoes a two-stage training process. In the first stage, the motion encoder-decoder (MED, \ie, $\mathcal{E}^M$ and $\mathcal{D}^M$) and the text encoder-decoder (TED, \ie, $\mathcal{E}^S$ and $\mathcal{D}^S$) are trained independently, we provide the ablation study about MED and TED in the supplement. Utilizing the trained encoders, we map motion sequences and textual descriptions into latent embeddings $z^m$ and $z^s$. In the second stage, our MDTM $\epsilon_{\theta}$ is trained to recover perturbed latent embeddings $z^m_{t_m}$ and $z^s_{t_s}$ from Gaussian noise $z^m_{T} \sim \mathcal{N}(0, I)$ and $z^s_{T} \sim \mathcal{N}(0, I)$. Here, we provide ablation studies for each component.

\begin{table*}[t]
    \centering
    \caption{Ablation study for the effectiveness of latent motion size in MTDM, i.e., the results with respect to $l$ for the text-to-motion task on the HumanML3D and KIT datasets.}
    \scalebox{1}{
    \begin{tabular}{llccccccc}
        \toprule
        \multirow{2}{*}{Dataset} &\multirow{2}{*}{Methods} & \multicolumn{3}{c}{R-Precision $\uparrow$} & \multirow{2}{*}{FID $\downarrow$} & \multirow{2}{*}{MMDist $\downarrow$} & \multirow{2}{*}{MModality $\uparrow$} \\ 
        \cline{3-5}
        & & Top1 & Top2 & Top3 &  &  &  &  \\ \hline 
        \multirow{5}{*}{HumanML3D}
        & Real &  $0.511^{\pm.003}$&  $0.703^{\pm.003}$&  $0.797^{\pm.002}$&  $0.002^{\pm.000}$&  $2.974^{\pm.008}$&  -\\\cline{2-9}
        & MTDM ($l$=2)  & $0.532^{\pm.003}$ & $0.723^{\pm.002}$ & $0.816^{\pm.002}$ & $0.191^{\pm.006}$ & $2.882^{\pm.007}$ & $2.060^{\pm.009}$	\\
        & MTDM ($l$=4)  & $\mathbf{0.548}^{\pm.002}$ & $\mathbf{0.737}^{\pm.002}$ & $\mathbf{0.825}^{\pm.002}$ & $\mathbf{0.075}^{\pm.004}$ & $\mathbf{2.867}^{\pm.012}$ & $2.399^{\pm.075}$	\\
        & MTDM ($l$=6)  & $0.504^{\pm.003}$ & $0.680^{\pm.003}$ & $0.765^{\pm.002}$ & $0.206^{\pm.010}$ & $3.285^{\pm.013}$ & $3.212^{\pm.111}$	\\
        & MTDM ($l$=8)  & $0.477^{\pm.002}$ & $0.653^{\pm.002}$ & $0.742^{\pm.002}$ & $0.213^{\pm.009}$ & $3.39^{\pm.013}$ & $\mathbf{3.286}^{\pm.103}$	\\\hline
        \multirow{5}{*}{KIT}
        & Real &  $0.424^{\pm.005}$&  $0.649^{\pm.006}$&  $0.779^{\pm.006}$&  $0.031^{\pm.004}$&  $2.788^{\pm.012}$&  -\\\cline{2-9} 
        & MTDM ($l$=2)  & $\mathbf{0.426}^{\pm.007}$& $\mathbf{0.640}^{\pm.005}$& $\mathbf{0.765}^{\pm.006}$& $0.404^{\pm.019}$& $\mathbf{2.905}^{\pm.025}$& $1.881^{\pm.078}$\\
        & MTDM ($l$=4)  & $0.419^{\pm.005}$& $0.627^{\pm.005}$& $0.741^{\pm.004}$& $0.256^{\pm.018}$& $3.316^{\pm.036}$& $\mathbf{2.615}^{\pm.103}$\\
        & MTDM ($l$=6)  & $0.394^{\pm.006}$& $0.604^{\pm.004}$& $0.727^{\pm.006}$& $\mathbf{0.213}^{\pm.012}$& $3.327^{\pm.043}$& $2.702^{\pm.094}$\\
        & MTDM ($l$=8)  & $0.372^{\pm.007}$& $0.573^{\pm.009}$& $0.696^{\pm.007}$& $0.434^{\pm.035}$& $3.602^{\pm.044}$& $3.125^{\pm.092}$\\
        \bottomrule
    \end{tabular}
    }
    \label{tab:ablation latent motion size in MTDM for text-to-motion}
\end{table*}

\begin{table*}[t]
    \centering
    \caption{Ablation study for the effectiveness of latent motion size in MTDM, i.e., the results with respect to $l$ for the motion-to-text task on the KIT dataset.}
    \scalebox{1}{
    \begin{tabular}{c c c c c c c c c}
        \toprule
        \multirow{2}{*}{Methods} & \multicolumn{2}{c}{R-Precision$\uparrow$} &\multirow{2}{*}{MMDist$\downarrow$} & \multirow{2}{*}{Bleu@1 $\uparrow$} & \multirow{2}{*}{Bleu@4 $\uparrow$} &  \multirow{2}{*}{Rouge $\uparrow$} &  \multirow{2}{*}{CIDEr $\uparrow$} & \multirow{2}{*}{BertScore $\uparrow$} \\ \cline{2-3}
        &  Top1 & Top3 &  &  &  &  &  & \\\hline
        Real & $0.422$ & $0.774$ & $2.775$ & - & - & - & - & - \\
        \cline{1-9} 
        MTDM ($l$=2) & $0.387$ & $0.724$ & $3.253$ & $44.8$ & $13.77$ &$40.9$ &$52.7$ &$33.3$ \\
        MTDM ($l$=4) & $\mathbf{0.421}$ & $\mathbf{0.765}$ & $\mathbf{2.976}$ & $44.9$ & $14.51$ &$41.8$ &$55.6$ &$\mathbf{35.9}$ \\
        MTDM ($l$=6) & $0.414$ & $0.758$ & $3.102$ & $46.2$ & $\mathbf{15.11}$ &$\mathbf{42.0}$ &$\mathbf{56.8}$ &$35.1$ \\
        MTDM ($l$=8) & $0.410$ & $0.747$ & $3.105$ & $\mathbf{46.3}$ & $15.10$ &$\mathbf{42.0}$ &$56.6$ &$34.8$\\
        \bottomrule
    \end{tabular}
    }
    \label{tab:ablation latent motion size in MTDM for motion-to-text}
\end{table*}

\noindent\textbf{Classifier-free Guidance.} In Tab.~\ref{tab:ablation cfg}, we conduct experiments on classifier-free guidance (CFG) for motion and text generation tasks. We find that $w_m=7.5$ achieves the best FID on the text-to-motion task, and $w_s=7.0$ performs best on the motion-to-text task, therefore we adopt $w_m=7.5$ and $w_s=7.0$ finally.

\begin{table*}[t]
\centering
    \caption{Classifier-free guidance with different $w_m$ and $w_s$ on the HumanML3D dataset.}
    \begin{tabular}{l c c c c c c}
    \toprule
        Text-to-Motion & R Top1$\uparrow$ &  R Top2$\uparrow$ &  R Top3$\uparrow$ & FID $\downarrow$ &MMDist $\downarrow$ & MModality $\uparrow$ \\\hline
         $w_m=0.0$ & $0.323^{\pm.003}$ & $0.467^{\pm.003}$ & $0.556^{\pm.003}$ & $0.986^{\pm.023}$ & $4.640^{\pm.018}$ & $\textbf{5.048}^{\pm.109}$ \\
         $w_m=3.0$ & $0.501^{\pm.003}$ & $0.685^{\pm.003}$ & $0.776^{\pm.002}$ & $0.146^{\pm.007}$ & $3.176^{\pm.012}$ & $3.009^{\pm.072}$ \\
         $w_m=7.5$ & $0.548^{\pm.002}$ & $0.737^{\pm.002}$ & $\textbf{0.825}^{\pm.002}$ & $\textbf{0.075}^{\pm.004}$ & $2.867^{\pm.012}$ & $2.399^{\pm.075}$	\\
         $w_m=8.0$ & $\textbf{0.550}^{\pm.003}$ & $\textbf{0.739}^{\pm.002}$ & $\textbf{0.825}^{\pm.002}$ & $0.106^{\pm.007}$ & $\textbf{2.861}^{\pm.006}$ & $2.278^{\pm.070}$ \\\hline
    Motion-to-Text & R Top3$\uparrow$  & MMDist$\downarrow$ & Bleu@4 $\uparrow$ & Rouge $\uparrow$& CIDEr $\uparrow$ & BertScore $\uparrow$ \\\hline
         $w_s=0.0$ & 0.610 & 4.198 & 6.34 & 29.9 & 14.9 & 20.4 \\
         $w_s=3.0$ & 0.821 & 2.869 & 10.07 & 35.9 & 27.8 & 28.7 \\
         $w_s=7.0$ & \textbf{0.871} & \textbf{2.649} & 11.15 & \textbf{37.4} & \textbf{31.5} & \textbf{30.3} \\
         $w_s=8.0$ & 0.851 & 2.720 & \textbf{11.25} & \textbf{37.4} & 31.0 & 29.9 \\
    \bottomrule
    \end{tabular}
    \label{tab:ablation cfg}
\end{table*}

\noindent\textbf{Effectiveness of Latent Motion Size in MTDM.} In Tab.~\ref{tab:ablation latent motion size in MTDM for text-to-motion}, we evaluate the impact of latent motion size $l$ on the performance of MTDM $\epsilon_{\theta}$ on the HumanML3D and KIT datasets. The same evaluation metrics in Tab.~\ref{tab:the state-of-the-art text-to-motion humanml3d} are reported. We find that $l=4$ and $l=2$ work best on the HumanML3D and KIT datasets, respectively. We observe that the text-to-motion performance decreases with an increase in $l$. Interestingly, the model with $l=2$ on KIT outperforms ``Real'' on R-Precision Top1, we guess this is caused by the biased text and motion feature extractor (similar results found in~\cite{guo2024momask}. In Tab.~\ref{tab:the state-of-the-art motion-to-text humanml3d} and Tab.~\ref{tab:ablation latent motion size in MTDM for motion-to-text}, we evaluate the motion-to-text performance of models with different latent motion sizes on two datasets. The experimental results demonstrate that a longer latent size $l$ achieves better motion-to-text performance in most metrics. We believe there exists a trade-off between the performance on text-to-motion and motion-to-text tasks; therefore, we finally select $l=4$ in the other experiments.

\begin{table*}[t]
    \centering
    \caption{Details of MoTe models. We name the models MoTe-S, MoTe-M, and MoTe-H, with an additional "-L" denoting a larger feedforward layer.}
    \scalebox{1}{
    \begin{tabular}{lccccc}
        \toprule
        \multirow{2}{*}{Model} &\multirow{2}{*}{\#DPD Block} &\multirow{2}{*}{Feedforward} & \multicolumn{3}{c}{Parameters(FLOPs)} \\ 
        \cline{4-6}
                   &                     &                  & In-Context  & Cross-Attention     & AdaLN \\ \hline 
         MoTe-S    &  7                  & 1024             & 94.9 M (2.256 G)      & 155 M (2.258 G)       & 188 M (4.335 G) \\
         MoTe-M    &  9                  & 1024             & 120 M (2.886 G)       & 198 M (2.888 G)       & 241 M (5.559 G)\\
         MoTe-H    &  11                 & 1024             & 146 M (3.515 G)      & 242 M (3.518 G)      & 294 M (6.782 G)\\
         MoTe-S-L  &  7                  & 3072             & 160 M (5.955 G)      & 243 M (5.957 G)      & 276 M (8.034 G)\\
         MoTe-M-L  &  9                  & 3072             & 205 M (7.642 G)      & 312 M (7.644 G)       & 354 M (10.315 G)\\
         MoTe-H-L  &  11                 & 3072             & 250 M (9.329 G)      & 380 M (9.331 G)       & 432 M (12.596 G)\\
        \bottomrule
    \end{tabular}
    }
    \label{tab:ablation_interaction}
\end{table*}

\begin{figure*}[t]
    \centering
    \subfloat[FID and Bleu@4 on the HumanML3D dataset.\label{fig:interaction humanml3d}]{%
        \includegraphics[width=0.4\textwidth]{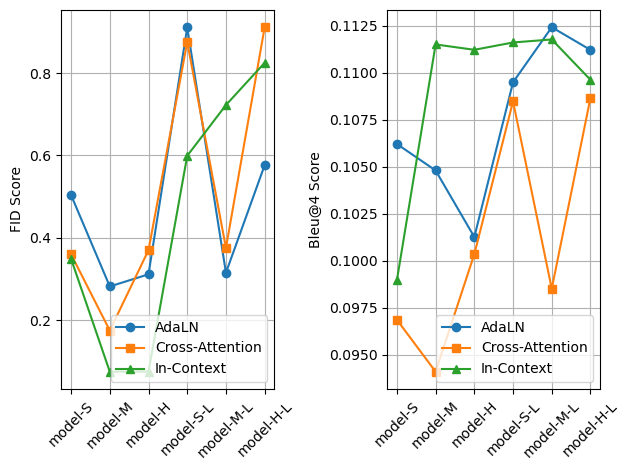}}
    \hfill
    \subfloat[Failure cases: (1) precise control; (2) out-of-domain description; (3) word repetition.\label{fig:interaction kit}]{%
        \includegraphics[width=0.6\linewidth]{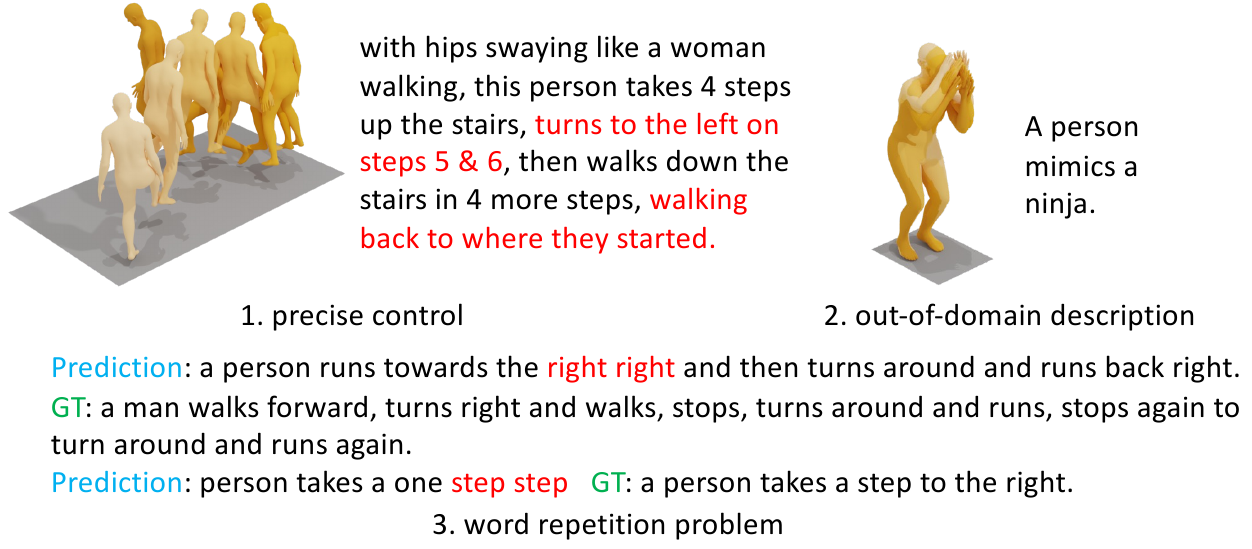}}
    
    \caption{(a) Comparison of different interaction modules at FID (lower is better) and Bleu@4 (higher is better) on the HumanML3D dataset. (b) Failure cases: 1. precise control; 2. out-of-domain description; 3. word repetition.}
    \label{fig:interaction}
\end{figure*}

\noindent\textbf{Effectiveness of Interaction Module in DPD Block.} As mentioned in Sec.~\ref{sec:mtdm}, the interaction module plays a pivot role in MTDM to empower the multimodal generation capability, we introduce three variants of interaction modules: In-Context Interaction, Cross-Attention Interaction, and AdaLN Interaction. We conduct an in-depth analysis of six variants of the MoTe model, each incorporating distinct self-attention layers and cross-modal interaction modules. As detailed in Tab.~\ref{tab:ablation_interaction}, we fix the input hidden dimension as 768 across all variants to ensure uniformity in our comparisons. The In-Context interaction module consistently exhibits lower parameter counts and flops compared to the Cross-Attention and AdaLN modules. Notably, the AdaLN module requires nearly double the parameters and flops than the In-Context module. We present the experimental results in Fig.~\ref{fig:interaction}. The In-Context module generally surpasses both AdaLN and Cross-Attention, particularly with smaller model sizes. %

\noindent\textbf{Remark.} In light of the comparison of different interaction modules, we observe two phenomena: 1)The performance of larger-sized models is not always better than their smaller-sized counterparts. We speculate that this discrepancy may be caused by the limited size of motion datasets. Interestingly, similar results were observed by the authors of MotionGPT~\cite{jiang2023motiongpt} when scaling the model size on the HumanML3D dataset.
2) The simplest In-Context interaction module achieves the best performance in most cases. This finding appears to contradict the conclusion of DiT, which suggests that AdaLN-Zero performs better than Cross-Attention and In-Context modules. As illustrated in Fig.~\ref{fig:interaction}, an individual AdaLN interaction module adopts two basic AdaLN modules, resulting in more parameters compared to the In-Context interaction module (around two times). However, according to the conclusion in DiT that Gflops is critical to improving performance, the In-Context interaction module obtains similar Gflops compared to the AdaLN interaction module when we fix the number of parameters. Therefore, the performance of the In-Context interaction module is close to that of the AdaLN interaction module.

\section{Conclusion}

In this work, we have proposed a unified diffusion model called MoTe to learn multiple distributions at the same time, therefore our method is able to handle diverse motion-related tasks with minor modification on the input context. Extensive experiments demonstrate that our method achieves superior text-to-motion performance and also achieves competitive performance for motion captioning when compared with the previous multi-modality generation works.

\noindent\textbf{Limitation and failure cases.} 
To delve into the limitation of MoTe, we present some detailed failure cases in Fig.~\ref{fig:interaction kit}. Our observations reveal specific limitations in our proposed method, for text-to-motion task: 1) \textbf{Precise Control}: Our model struggles with accurately capturing fine-grained descriptions for precise motion control. Implementing a coarse-to-fine text feature extraction approach could potentially mitigate this issue. 2) \textbf{Out-of-Domain Descriptions}: The model faces challenges in adapting to out-of-domain textual descriptions. A potential solution to enhance its adaptability could be the development and utilization of larger and more diverse training datasets. For motion-to-text task: \textbf{Word Repetition Problem}. Despite exploring various sampling methods, our model has not fully overcome the repetitive text generation issue, a limitation inherited from the underlying GPT2 architecture. We believe that integrating a more advanced text generation model (e.g. T5) may effectively resolve this problem. Except for the observed failure cases, our work aims to address the multi-modal generation problem for a single character, which limits its application in complex scenario. How to extend our method for human-object and human-human generation is our future work.

\bibliographystyle{IEEEtran}
\bibliography{IEEEabrv,main}

\begin{thebibliography}{10}
\providecommand{\url}[1]{#1}
\csname url@samestyle\endcsname
\providecommand{\newblock}{\relax}
\providecommand{\bibinfo}[2]{#2}
\providecommand{\BIBentrySTDinterwordspacing}{\spaceskip=0pt\relax}
\providecommand{\BIBentryALTinterwordstretchfactor}{4}
\providecommand{\BIBentryALTinterwordspacing}{\spaceskip=\fontdimen2\font plus
\BIBentryALTinterwordstretchfactor\fontdimen3\font minus \fontdimen4\font\relax}
\providecommand{\BIBforeignlanguage}[2]{{%
\expandafter\ifx\csname l@#1\endcsname\relax
\typeout{** WARNING: IEEEtran.bst: No hyphenation pattern has been}%
\typeout{** loaded for the language `#1'. Using the pattern for}%
\typeout{** the default language instead.}%
\else
\language=\csname l@#1\endcsname
\fi
#2}}
\providecommand{\BIBdecl}{\relax}
\BIBdecl

\bibitem{rombach2022high}
R.~Rombach, A.~Blattmann, D.~Lorenz, P.~Esser, and B.~Ommer, ``High-resolution image synthesis with latent diffusion models,'' in \emph{CVPR}.\hskip 1em plus 0.5em minus 0.4em\relax {IEEE}, 2022, pp. 10\,674--10\,685.

\bibitem{poole2023dreamfusion}
B.~Poole, A.~Jain, J.~T. Barron, and B.~Mildenhall, ``Dreamfusion: Text-to-3d using 2d diffusion,'' in \emph{ICLR}, 2023.

\bibitem{ho2020denosing}
J.~Ho, A.~Jain, and P.~Abbeel, ``Denoising diffusion probabilistic models,'' in \emph{NeurIPS}, H.~Larochelle, M.~Ranzato, R.~Hadsell, M.~Balcan, and H.~Lin, Eds., 2020.

\bibitem{song19generative}
Y.~Song and S.~Ermon, ``Generative modeling by estimating gradients of the data distribution,'' in \emph{NeurIPS}, H.~M. Wallach, H.~Larochelle, A.~Beygelzimer, F.~d'Alch{\'{e}}{-}Buc, E.~B. Fox, and R.~Garnett, Eds., 2019, pp. 11\,895--11\,907.

\bibitem{radford2021learning}
A.~Radford, J.~W. Kim, C.~Hallacy, A.~Ramesh, G.~Goh, S.~Agarwal, G.~Sastry, A.~Askell, P.~Mishkin, J.~Clark, G.~Krueger, and I.~Sutskever, ``Learning transferable visual models from natural language supervision,'' in \emph{ICML}, ser. Proceedings of Machine Learning Research, M.~Meila and T.~Zhang, Eds., vol. 139.\hskip 1em plus 0.5em minus 0.4em\relax {PMLR}, 2021, pp. 8748--8763.

\bibitem{tevet2022human}
G.~Tevet, S.~Raab, B.~Gordon, Y.~Shafir, D.~Cohen-or, and A.~H. Bermano, ``Human motion diffusion model,'' in \emph{ICLR}, 2022.

\bibitem{zhang2022motiondiffuse}
M.~Zhang, Z.~Cai, L.~Pan, F.~Hong, X.~Guo, L.~Yang, and Z.~Liu, ``Motiondiffuse: Text-driven human motion generation with diffusion model,'' \emph{arXiv preprint arXiv:2208.15001}, 2022.

\bibitem{chen2023executing}
X.~Chen, B.~Jiang, W.~Liu, Z.~Huang, B.~Fu, T.~Chen, and G.~Yu, ``Executing your commands via motion diffusion in latent space,'' in \emph{CVPR}, 2023, pp. 18\,000--18\,010.

\bibitem{zhang2023t2mgpt}
J.~Zhang, Y.~Zhang, X.~Cun, S.~Huang, Y.~Zhang, H.~Zhao, H.~Lu, and X.~Shen, ``T2m-gpt: Generating human motion from textual descriptions with discrete representations,'' \emph{CVPR}, 2023.

\bibitem{guo2022tm2t}
C.~Guo, X.~Zuo, S.~Wang, and L.~Cheng, ``Tm2t: Stochastic and tokenized modeling for the reciprocal generation of 3d human motions and texts,'' in \emph{ECCV}.\hskip 1em plus 0.5em minus 0.4em\relax Springer, 2022, pp. 580--597.

\bibitem{jiang2023motiongpt}
B.~Jiang, X.~Chen, W.~Liu, J.~Yu, G.~Yu, and T.~Chen, ``Motiongpt: Human motion as a foreign language,'' \emph{NeurIPS}, 2023.

\bibitem{petrovich2021action}
M.~Petrovich, M.~J. Black, and G.~Varol, ``Action-conditioned 3d human motion synthesis with transformer {VAE},'' in \emph{ICCV}.\hskip 1em plus 0.5em minus 0.4em\relax {IEEE}, 2021, pp. 10\,965--10\,975.

\bibitem{guo2022generating}
C.~Guo, S.~Zou, X.~Zuo, S.~Wang, W.~Ji, X.~Li, and L.~Cheng, ``Generating diverse and natural 3d human motions from text,'' in \emph{CVPR}, 2022, pp. 5152--5161.

\bibitem{li2021ai}
R.~Li, S.~Yang, D.~A. Ross, and A.~Kanazawa, ``Ai choreographer: Music conditioned 3d dance generation with aist++,'' in \emph{ICCV}, 2021, pp. 13\,401--13\,412.

\bibitem{plappert2018learning}
M.~Plappert, C.~Mandery, and T.~Asfour, ``Learning a bidirectional mapping between human whole-body motion and natural language using deep recurrent neural networks,'' \emph{Robotics and Autonomous Systems}, vol. 109, pp. 13--26, 2018.

\bibitem{petrovich2022temos}
M.~Petrovich, M.~J. Black, and G.~Varol, ``Temos: Generating diverse human motions from textual descriptions,'' in \emph{ECCV}.\hskip 1em plus 0.5em minus 0.4em\relax Springer, 2022, pp. 480--497.

\bibitem{goutsu2021linguistic}
Y.~Goutsu and T.~Inamura, ``Linguistic descriptions of human motion with generative adversarial seq2seq learning,'' in \emph{{IEEE} International Conference on Robotics and Automation, {ICRA} 2021, Xi'an, China, May 30 - June 5, 2021}.\hskip 1em plus 0.5em minus 0.4em\relax {IEEE}, 2021, pp. 4281--4287.

\bibitem{wang2023fg}
Y.~Wang, Z.~Leng, F.~W. Li, S.-C. Wu, and X.~Liang, ``Fg-t2m: Fine-grained text-driven human motion generation via diffusion model,'' in \emph{ICCV}, 2023, pp. 22\,035--22\,044.

\bibitem{yamada2018paired}
T.~Yamada, H.~Matsunaga, and T.~Ogata, ``Paired recurrent autoencoders for bidirectional translation between robot actions and linguistic descriptions,'' \emph{{IEEE} Robotics Autom. Lett.}, vol.~3, no.~4, pp. 3441--3448, 2018.

\bibitem{chung2022scaling}
H.~W. Chung, L.~Hou, S.~Longpre, B.~Zoph, Y.~Tay, W.~Fedus, Y.~Li, X.~Wang, M.~Dehghani, S.~Brahma \emph{et~al.}, ``Scaling instruction-finetuned language models,'' \emph{arXiv preprint arXiv:2210.11416}, 2022.

\bibitem{touvron2023llama}
H.~Touvron, T.~Lavril, G.~Izacard, X.~Martinet, M.-A. Lachaux, T.~Lacroix, B.~Rozi{\`e}re, N.~Goyal, E.~Hambro, F.~Azhar \emph{et~al.}, ``Llama: Open and efficient foundation language models,'' \emph{arXiv preprint arXiv:2302.13971}, 2023.

\bibitem{cai2023smpler}
Z.~Cai, W.~Yin, A.~Zeng, C.~Wei, Q.~Sun, Y.~Wang, H.~E. Pang, H.~Mei, M.~Zhang, L.~Zhang \emph{et~al.}, ``Smpler-x: Scaling up expressive human pose and shape estimation,'' \emph{arXiv preprint arXiv:2309.17448}, 2023.

\bibitem{wang2022image}
W.~Wang, H.~Bao, L.~Dong, J.~Bjorck, Z.~Peng, Q.~Liu, K.~Aggarwal, O.~K. Mohammed, S.~Singhal, S.~Som \emph{et~al.}, ``Image as a foreign language: Beit pretraining for all vision and vision-language tasks,'' \emph{arXiv preprint arXiv:2208.10442}, 2022.

\bibitem{girdhar2023imagebind}
R.~Girdhar, A.~El-Nouby, Z.~Liu, M.~Singh, K.~V. Alwala, A.~Joulin, and I.~Misra, ``Imagebind: One embedding space to bind them all,'' in \emph{CVPR}, 2023, pp. 15\,180--15\,190.

\bibitem{ruan2023mmdiffusion}
L.~Ruan, Y.~Ma, H.~Yang, H.~He, B.~Liu, J.~Fu, N.~J. Yuan, Q.~Jin, and B.~Guo, ``Mm-diffusion: Learning multi-modal diffusion models for joint audio and video generation,'' in \emph{CVPR}.\hskip 1em plus 0.5em minus 0.4em\relax {IEEE}, 2023, pp. 10\,219--10\,228.

\bibitem{bao2023one}
F.~Bao, S.~Nie, K.~Xue, C.~Li, S.~Pu, Y.~Wang, G.~Yue, Y.~Cao, H.~Su, and J.~Zhu, ``One transformer fits all distributions in multi-modal diffusion at scale,'' in \emph{ICML}, ser. Proceedings of Machine Learning Research, A.~Krause, E.~Brunskill, K.~Cho, B.~Engelhardt, S.~Sabato, and J.~Scarlett, Eds., vol. 202.\hskip 1em plus 0.5em minus 0.4em\relax {PMLR}, 2023, pp. 1692--1717.

\bibitem{alayrac2022flamingo}
J.-B. Alayrac, J.~Donahue, P.~Luc, A.~Miech, I.~Barr, Y.~Hasson, K.~Lenc, A.~Mensch, K.~Millican, M.~Reynolds \emph{et~al.}, ``Flamingo: a visual language model for few-shot learning,'' in \emph{NeurIPS}, vol.~35, 2022, pp. 23\,716--23\,736.

\bibitem{li2023blip}
J.~Li, D.~Li, S.~Savarese, and S.~Hoi, ``Blip-2: Bootstrapping language-image pre-training with frozen image encoders and large language models,'' \emph{arXiv preprint arXiv:2301.12597}, 2023.

\bibitem{liu2023visual}
H.~Liu, C.~Li, Q.~Wu, and Y.~J. Lee, ``Visual instruction tuning,'' \emph{arXiv preprint arXiv:2304.08485}, 2023.

\bibitem{nichol2021improved}
A.~Q. Nichol and P.~Dhariwal, ``Improved denoising diffusion probabilistic models,'' in \emph{ICML}.\hskip 1em plus 0.5em minus 0.4em\relax PMLR, 2021, pp. 8162--8171.

\bibitem{tevet22motionclip}
G.~Tevet, B.~Gordon, A.~Hertz, A.~H. Bermano, and D.~Cohen{-}Or, ``Motionclip: Exposing human motion generation to {CLIP} space,'' in \emph{ECCV}, ser. Lecture Notes in Computer Science, S.~Avidan, G.~J. Brostow, M.~Ciss{\'{e}}, G.~M. Farinella, and T.~Hassner, Eds., vol. 13682.\hskip 1em plus 0.5em minus 0.4em\relax Springer, 2022, pp. 358--374.

\bibitem{peebles2023scalable}
W.~Peebles and S.~Xie, ``Scalable diffusion models with transformers,'' in \emph{ICCV}, 2023, pp. 4195--4205.

\bibitem{vaswani2017attention}
A.~Vaswani, N.~Shazeer, N.~Parmar, J.~Uszkoreit, L.~Jones, A.~N. Gomez, L.~Kaiser, and I.~Polosukhin, ``Attention is all you need,'' in \emph{NeurIPS}, I.~Guyon, U.~von Luxburg, S.~Bengio, H.~M. Wallach, R.~Fergus, S.~V.~N. Vishwanathan, and R.~Garnett, Eds., 2017, pp. 5998--6008.

\bibitem{karras2021astyle}
T.~Karras, S.~Laine, and T.~Aila, ``A style-based generator architecture for generative adversarial networks,'' \emph{IEEE TPAMI}, vol.~43, no.~12, pp. 4217--4228, 2021.

\bibitem{ho2021classifier}
J.~Ho and T.~Salimans, ``Classifier-free diffusion guidance,'' in \emph{NeurIPS 2021 Workshop on Deep Generative Models and Downstream Applications}, 2021.

\bibitem{papineni2002bleu}
K.~Papineni, S.~Roukos, T.~Ward, and W.~Zhu, ``Bleu: a method for automatic evaluation of machine translation,'' in \emph{Association for Computational Linguistics.}, 2002, pp. 311--318.

\bibitem{lin2004rouge}
C.-Y. Lin, ``Rouge: A package for automatic evaluation of summaries,'' in \emph{Text summarization branches out}, 2004, pp. 74--81.

\bibitem{vedantam2015cider}
R.~Vedantam, C.~L. Zitnick, and D.~Parikh, ``Cider: Consensus-based image description evaluation,'' in \emph{CVPR}.\hskip 1em plus 0.5em minus 0.4em\relax {IEEE} Computer Society, 2015, pp. 4566--4575.

\bibitem{zhang2020bertscore}
T.~Zhang, V.~Kishore, F.~Wu, K.~Q. Weinberger, and Y.~Artzi, ``Bertscore: Evaluating text generation with {BERT},'' in \emph{ICLR}, 2020.

\bibitem{loshchilov2017sgdr}
I.~Loshchilov and F.~Hutter, ``{SGDR:} stochastic gradient descent with warm restarts,'' in \emph{ICLR}, 2017.

\bibitem{han2024hutumotion}
G.~Han, S.~Huang, M.~Gong, and J.~Tang, ``Hutumotion: Human-tuned navigation of latent motion diffusion models with minimal feedback,'' in \emph{AAAI}, 2024.

\bibitem{zhang2023remodiffuse}
\BIBentryALTinterwordspacing
M.~Zhang, X.~Guo, L.~Pan, Z.~Cai, F.~Hong, H.~Li, L.~Yang, and Z.~Liu, ``Remodiffuse: Retrieval-augmented motion diffusion model,'' in \emph{ICCV}.\hskip 1em plus 0.5em minus 0.4em\relax Los Alamitos, CA, USA: IEEE Computer Society, oct 2023, pp. 364--373. [Online]. Available: \url{https://doi.ieeecomputersociety.org/10.1109/ICCV51070.2023.00040}
\BIBentrySTDinterwordspacing

\bibitem{cai2024digital}
Z.~Cai, J.~Jiang, Z.~Qing, X.~Guo, M.~Zhang, Z.~Lin, H.~Mei, C.~Wei, R.~Wang, W.~Yin \emph{et~al.}, ``Digital life project: Autonomous 3d characters with social intelligence,'' in \emph{CVPR}, 2024, pp. 582--592.

\bibitem{guo2024momask}
C.~Guo, Y.~Mu, M.~G. Javed, S.~Wang, and L.~Cheng, ``Momask: Generative masked modeling of 3d human motions,'' in \emph{CVPR}, 2024.

\end{thebibliography}

\end{document}